\documentclass[letterpaper, 10 pt, conference]{ieeeconf}  

\usepackage[dvips]{graphicx}
\usepackage{bm}
\usepackage{cite}
\usepackage{flushend}
\include{preamble}

\IEEEoverridecommandlockouts                              

\title{\LARGE \textbf
  {
    \switchlanguage%
    {%
			WARL: Wrench-Augmented Reinforcement Learning for Task-Agnostic Learning in Legged Robots
		}%
		{%
			WARL: Wrench-Augmented Reinforcement Learning for Task-Agnostic Learning in Legged Robots
		}%
  }
}

\author{Keita Yoneda$^{1}$, Kento Kawaharazuka$^{1, 2}$, Kei Okada$^{1}$
  \thanks{$^{1}$ The authors are with the Department of Mechano-Informatics, Graduate School of Information Science and Technology, The University of Tokyo, 7-3-1 Hongo, Bunkyo-ku, Tokyo, 113-8656, Japan.
    {\texttt\small [yoneda, kawaharazuka, k-okada]@jsk.imi.i.u-tokyo.ac.jp}
  }
	\thanks{$^{2}$ The author is with the AI Center, Graduate School of Information Science and Technology, The University of Tokyo, Japan.
	}
}
\begin{document}

\maketitle
\thispagestyle{empty}
\pagestyle{empty}

\begin{abstract}
\switchlanguage%
{
	While reinforcement learning for legged robots has achieved high motor performance, it has been constrained by the limited exploration capability of actions confined to the joint space. To address this issue, this study proposes a new method, Wrench-Augmented Reinforcement Learning (WARL), which introduces a wrenche (force and torque) into the action space. The proposed method combines wrench-guided exploration with a success rate-based curriculum mechanism to expand exploration capabilities in the early stages of learning, with the ultimate goal of acquiring behaviors based solely on joint control.
	Experiments using a quadruped robot demonstrated that WARL can learn robustly across diverse terrains and motor tasks without requiring terrain-specific reward adjustments or complex curriculum designs. Furthermore, an ablation study verified the effectiveness of the Switching Curriculum, which gradually eliminates the wrench.
	On the other hand, we also show that introducing a wrench can encourage behaviors that do not sufficiently exploit the robot's physical embodiment. These findings suggest that while wrench-based exploration enhancement is effective for improving learning efficiency, designing it in a way that is consistent with the robot's physical structure is a critical future challenge.
}
{
	è„šãƒ­ãƒœãƒƒãƒˆã�®å¼·åŒ–å­¦ç¿’ã�¯é«˜ã�„é�‹å‹•æ€§èƒ½ã‚’å®Ÿç�¾ã�™ã‚‹ä¸€æ–¹,é–¢ç¯€ç©ºé–“ã�«é™�å®šã�•ã‚Œã�Ÿã‚¢ã‚¯ã‚·ãƒ§ãƒ³ã�«ã‚ˆã‚‹æŽ¢ç´¢èƒ½åŠ›ã�®åˆ¶ç´„ã�Œèª²é¡Œã�§ã�‚ã�£ã�Ÿ.æœ¬ç ”ç©¶ã�§ã�¯ã�“ã�®èª²é¡Œã�«å¯¾ã�—,ã‚¢ã‚¯ã‚·ãƒ§ãƒ³ç©ºé–“ã�«ãƒ¬ãƒ³ãƒ�(åŠ›ã�¨ãƒˆãƒ«ã‚¯)ã‚’å°Žå…¥ã�™ã‚‹æ–°æ‰‹æ³•Wrench-Augmented Reinforcement Learning(WARL)ã‚’æ��æ¡ˆã�™ã‚‹.æ��æ¡ˆæ‰‹æ³•ã�¯,ãƒ¬ãƒ³ãƒ�ã�«ã‚ˆã‚‹æŽ¢ç´¢èª˜å°Žã�¨æˆ�åŠŸçŽ‡ã�«åŸºã�¥ã��ã‚«ãƒªã‚­ãƒ¥ãƒ©ãƒ æ©Ÿæ§‹ã‚’çµ„ã�¿å�ˆã‚�ã�›ã‚‹ã�“ã�¨ã�§,å­¦ç¿’åˆ�æœŸã�®æŽ¢ç´¢èƒ½åŠ›ã‚’æ‹¡å¼µã�—,æœ€çµ‚çš„ã�«ã�¯é–¢ç¯€åˆ¶å¾¡ã�®ã�¿ã�«ã‚ˆã‚‹å‹•ä½œç�²å¾—ã‚’ç›®æŒ‡ã�™.

å››è„šãƒ­ãƒœãƒƒãƒˆã‚’ç”¨ã�„ã�Ÿå®Ÿé¨“ã�«ã‚ˆã‚Š,WARLã�Œå¤šæ§˜ã�ªåœ°å½¢ã�Šã‚ˆã�³é�‹å‹•èª²é¡Œã�«å¯¾ã�—,åœ°å½¢å›ºæœ‰ã�®å ±é…¬èª¿æ•´ã‚„è¤‡é›‘ã�ªã‚«ãƒªã‚­ãƒ¥ãƒ©ãƒ è¨­è¨ˆã‚’å¿…è¦�ã�¨ã�›ã�š,çµ±ä¸€çš„ã�«å­¦ç¿’å�¯èƒ½ã�§ã�‚ã‚‹ã�“ã�¨ã‚’å®Ÿè¨¼ã�—ã�Ÿ.ã�•ã‚‰ã�«,ãƒ¬ãƒ³ãƒ�ã‚’æ®µéšŽçš„ã�«é™¤åŽ»ã�™ã‚‹Switching Curriculumã�®æœ‰åŠ¹æ€§ã‚’ã‚¢ãƒ–ãƒ¬ãƒ¼ã‚·ãƒ§ãƒ³ã‚¹ã‚¿ãƒ‡ã‚£ã�«ã‚ˆã�£ã�¦æ¤œè¨¼ã�—ã�Ÿ.

ä¸€æ–¹ã�§,ãƒ¬ãƒ³ãƒ�ã�®å°Žå…¥ã�Œèº«ä½“æ€§ã‚’å��åˆ†ã�«è€ƒæ…®ã�—ã�ªã�„å‹•ä½œã�®å­¦ç¿’ã‚’ä¿ƒã�™å�¯èƒ½æ€§ã‚‚ç¤ºå”†ã�•ã‚Œã�Ÿ.ã�“ã‚Œã‚‰ã�®çµ�æžœã�¯,ãƒ¬ãƒ³ãƒ�ã‚’ç”¨ã�„ã�ŸæŽ¢ç´¢æ‹¡å¼µã�Œå­¦ç¿’åŠ¹çŽ‡ã�®å�‘ä¸Šã�«æœ‰åŠ¹ã�§ã�‚ã‚‹ä¸€æ–¹,èº«ä½“æ§‹é€ ã�¨ã�®æ•´å�ˆæ€§ã‚’è€ƒæ…®ã�—ã�Ÿè¨­è¨ˆã�Œä»Šå¾Œã�®é‡�è¦�ã�ªèª²é¡Œã�§ã�‚ã‚‹ã�“ã�¨ã‚’ç¤ºå”†ã�™ã‚‹.
}
\end{abstract}

\section{Introduction}\label{sec:introduction}
\switchlanguage%
{%
Legged robots have garnered attention as highly mobile platforms capable of traversing diverse environments.
In recent years, reinforcement learning has become a standard method for achieving dynamic motions such as walking on uneven terrain and jumping \cite{hwangbo2019anymal, zhuang2023robot, kim2025high}.
However, conventional frameworks often involve numerous heuristics in the learning environment and reward design for specific tasks, posing a challenge due to the high design cost for new tasks.

For example, in tasks that involve traversing uneven terrain with large steps, a standard approach is to use a heuristic curriculum design that gradually increases the height of the steps as learning progresses \cite{rudin2022leggedgym}.
Even when not using curriculum-based changes in terrain, task-specific, meticulous reward design, such as reward shaping that incorporates fine-grained behaviors for learning jumping motions, is often necessary \cite{atanassov2024curriculum}.
While these methods have been successful in learning specific tasks, they lack the versatility to be applied to a wide range of tasks.

\begin{figure}[tbp]
    \centering
        \includegraphics[width=0.95\columnwidth]{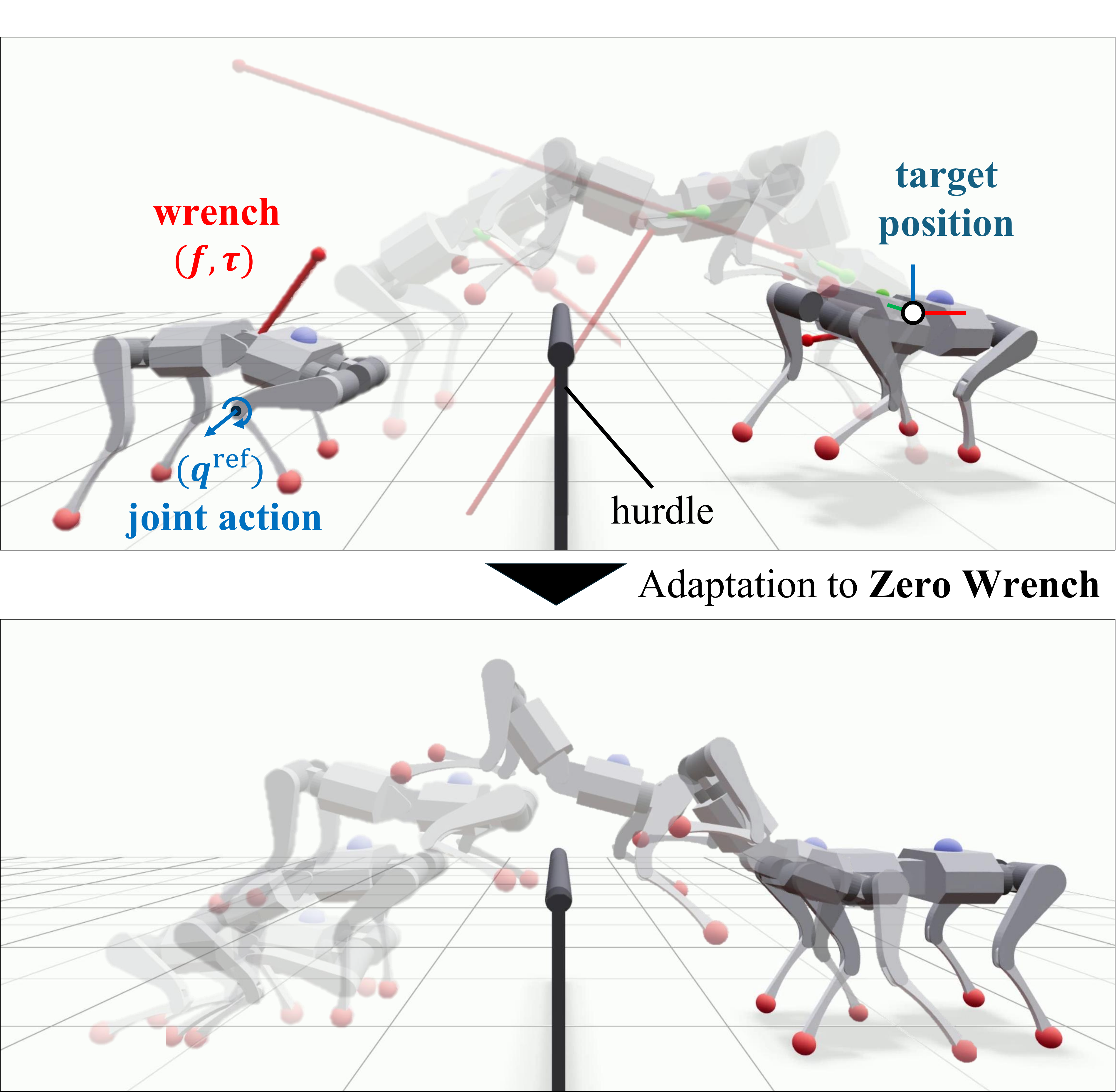}
\caption{Overview of Wrench-Augmented Reinforcement Learning (WARL). By introducing a wrench into the action space, exploration is promoted, and a policy that does not use the wrench is ultimately acquired through a curriculum learning process.}
\label{fig:warl_overview}
\vspace{-3ex}
\end{figure}

In this study, we hypothesize that a primary reason for the need for task-specific design is the difficulty of exploration, which stems from the complex relationship between the action space and the task space.
Here, the task space refers to the space that characterizes the motion of the legged robot, which, for locomotion tasks, includes the robot's center of mass position and orientation.
In conventional reinforcement learning for legged robots, the action space is typically the joint space, such as joint angles or joint torques \cite{margolis2024rapid, ha2025learning}.
In general, the mapping from the joint space to the task space is complex because it is strongly influenced by discrete and nonlinear factors such as contact states and joint configurations.
This complexity can lead to slow learning speeds and convergence to local optima.
Previous research has attempted to address this problem through modifications in environment and reward design \cite{vogel2024robust_ladder,bellegarda2024robust,yoneda2025kleiyn}.

However, a more direct solution is to reduce the mismatch between the action and task spaces.
For example, in robot arm control, it has been reported that using the end-effector position as the action improves exploration efficiency \cite{martin2019variable, jiang2024learning, aljalbout2024role}.
For legged robots, methods have been proposed that use the foot-end space as the action \cite{duan2021learning}, hierarchical methods that combine model-based control with high-level actions such as foot-end position and gait phase \cite{bellegarda2022cpg, kang2025dynamic}, and methods that introduce prior structures such as periodic trajectories \cite{miki2022anymal}.
Although these methods contribute to improving learning efficiency, there are limits to the improvement in exploration performance because nonlinearity still exists between the foot-end space and the robot's body position and orientation space.
Furthermore, design constraints for real-world deployment often limit the choice of action parameterization.

One of the essential features of legged robot locomotion is the change in the robot's position and orientation. By incorporating a wrench ($w=(f,\tau)\in\mathbb{R}^6$), which directly acts on these quantities, into the action space, the relationship to the task space can be simplified, thereby improving exploration. Previous studies have introduced wrenches into motor learning, including EFGCL, which applies external forces to facilitate backflip, barrel roll, and jumping motions\cite{yoneda2026efgcl}, ZEST, which uses model-based wrench generation for reference trajectory tracking\cite{sleiman2026zest}, and A2CF, which outputs wrenches to accelerate learning\cite{cao2025learning}. However, EFGCL relies on manually designed external forces requiring heuristic intuition, ZEST does not use wrenches for exploration, and A2CF employs them mainly as auxiliary assistance rather than as exploratory actions. Moreover, evaluations have been limited to tasks solvable without wrench assistance, leaving the effectiveness of wrench-guided exploration insufficiently validated.

To address these limitations, we propose a method that introduces a wrench as a primary exploratory action and gradually removes its influence through a curriculum learning scheme. During early training, behavior is generated mainly through the wrench, whose contribution is progressively reduced as learning proceeds. By directly applying policy-generated wrenches to the robot's torso, the proposed method promotes efficient exploration in locomotion tasks while ultimately learning an equivalent joint-only control policy.

The main contributions of this study are as follows:
\begin{enumerate}
\item We show that by introducing a wrench into the action space in the motor learning of legged robots, exploration can be made more efficient, and versatile learning that does not heavily depend on task-specific environment or reward design becomes possible.
\item We propose a curriculum learning method for acquiring a policy that can be applied to real robots by gradually removing the wrench introduced for exploration guidance.
\item We analyze that there is a trade-off between the improvement in spatial exploration ability by this method and the difficulty in acquiring dexterous motions that utilize the environment, such as footholds. Resolving this trade-off is a future challenge.
\end{enumerate}
}%
{%
è„šãƒ­ãƒœãƒƒãƒˆã�¯ã��ã�®é«˜ã�„æ©Ÿå‹•æ€§ã�‹ã‚‰,å¤šæ§˜ã�ªç’°å¢ƒã‚’è¸�ç ´å�¯èƒ½ã�ªãƒ—ãƒ©ãƒƒãƒˆãƒ•ã‚©ãƒ¼ãƒ ã�¨ã�—ã�¦æ³¨ç›®ã�•ã‚Œã�¦ã�„ã‚‹.
è¿‘å¹´ã�§ã�¯ä¸�æ•´åœ°æ­©è¡Œã‚„è·³èº�ã�¨ã�„ã�£ã�Ÿå‹•çš„é�‹å‹•ã‚’å®Ÿç�¾ã�™ã‚‹æ‰‹æ³•ã�¨ã�—ã�¦å¼·åŒ–å­¦ç¿’ã‚’ç”¨ã�„ã‚‹ã�“ã�¨ã�Œæ¨™æº–ã�¨ã�ªã‚Šã�¤ã�¤ã�‚ã‚‹\cite{hwangbo2019anymal, zhuang2023robot, kim2025high}.
ã�—ã�‹ã�—,å¾“æ�¥ã�®ãƒ•ãƒ¬ãƒ¼ãƒ ãƒ¯ãƒ¼ã‚¯ã�§ã�¯,ç‰¹å®šã�®ã‚¿ã‚¹ã‚¯ã�®å­¦ç¿’ç’°å¢ƒã‚„å ±é…¬è¨­è¨ˆã�«å¤šã��ã�®ãƒ’ãƒ¥ãƒ¼ãƒªã‚¹ãƒ†ã‚£ãƒƒã‚¯ã�Œå�«ã�¾ã‚Œã�¦ã�Šã‚Š,æ–°è¦�ã‚¿ã‚¹ã‚¯ã�«å¯¾ã�™ã‚‹è¨­è¨ˆã‚³ã‚¹ãƒˆã�®é«˜ã�•ã�Œèª²é¡Œã�§ã�‚ã�£ã�Ÿ.

ä¾‹ã�ˆã�°,å¤§ã��ã�ªæ®µå·®ã‚’å�«ã‚€ä¸�æ•´åœ°ã‚’èµ°ç ´ã�™ã‚‹ã‚¿ã‚¹ã‚¯ã�§ã�¯,å­¦ç¿’ã�®é€²è¡Œã�«å¿œã�˜ã�¦æ®µå·®ã‚’é«˜ã��ã�—ã�¦ã�„ã��ã�¨ã�„ã�£ã�Ÿãƒ’ãƒ¥ãƒ¼ãƒªã‚¹ãƒ†ã‚£ãƒƒã‚¯ã�ªã‚«ãƒªã‚­ãƒ¥ãƒ©ãƒ è¨­è¨ˆã�Œæ¨™æº–çš„ã�«ç”¨ã�„ã‚‰ã‚Œã‚‹\cite{rudin2022leggedgym}.
ã�¾ã�Ÿ,åœ°å½¢ã�®ã‚«ãƒªã‚­ãƒ¥ãƒ©ãƒ çš„å¤‰åŒ–ã‚’åˆ©ç”¨ã�—ã�ªã�„å ´å�ˆã�§ã‚‚,ã‚¸ãƒ£ãƒ³ãƒ—å‹•ä½œã�®å­¦ç¿’ã�®ã�Ÿã‚�ã�«ç´°ã�‹ã�„æŒ™å‹•ã‚’å ±é…¬ã�«çµ„ã�¿è¾¼ã‚€Reward-Shapingã�ªã�©,ã‚¿ã‚¹ã‚¯å›ºæœ‰ã�®å…¥å¿µã�ªå ±é…¬è¨­è¨ˆã�Œå¿…è¦�ã�¨ã�ªã‚‹ã�“ã�¨ã�Œå¤šã�„\cite{atanassov2024curriculum}.
ã�“ã‚Œã‚‰ã�®æ–¹æ³•ã�¯ç‰¹å®šã�®ã‚¿ã‚¹ã‚¯ã�®å­¦ç¿’ã�«ã�¯æˆ�åŠŸã�—ã�¦ã�„ã‚‹ã�Œ,å¤šæ§˜ã�ªã‚¿ã‚¹ã‚¯ã�«é�©ç”¨ã�§ã��ã‚‹æ±Žç”¨æ€§ã�«æ¬ ã�‘ã‚‹ã�¨ã�„ã�†å•�é¡Œã�Œã�‚ã�£ã�Ÿ.

æœ¬ç ”ç©¶ã�§ã�¯,ã‚¿ã‚¹ã‚¯å›ºæœ‰ã�®è¨­è¨ˆã‚’è¦�ã�™ã‚‹ä¸€å› ã�¯,ã‚¢ã‚¯ã‚·ãƒ§ãƒ³ç©ºé–“ã�¨ã‚¿ã‚¹ã‚¯ç©ºé–“ã�®é–“ã�®è¤‡é›‘ã�ªå¯¾å¿œé–¢ä¿‚ã�«èµ·å› ã�™ã‚‹,æŽ¢ç´¢ã�®å›°é›£æ€§ã�«ã�‚ã‚‹ã�¨æˆ‘ã€…ã�¯è€ƒã�ˆã‚‹.
ã�“ã�“ã�§ã‚¿ã‚¹ã‚¯ç©ºé–“ã�¨ã�¯è„šãƒ­ãƒœãƒƒãƒˆã�®é�‹å‹•ã‚’ç‰¹å¾´ä»˜ã�‘ã‚‹ç©ºé–“ã‚’æŒ‡ã�—,ç§»å‹•ã‚¿ã‚¹ã‚¯ã�§ã�¯ãƒ­ãƒœãƒƒãƒˆã�®ä¸­å¿ƒä½�ç½®ã‚„å§¿å‹¢ã�ªã�©ã�Œè©²å½“ã�™ã‚‹.
å¾“æ�¥ã�®è„šãƒ­ãƒœãƒƒãƒˆã�®å¼·åŒ–å­¦ç¿’ã�§ã�¯,ã‚¢ã‚¯ã‚·ãƒ§ãƒ³ç©ºé–“ã�¨ã�—ã�¦é–¢ç¯€è§’åº¦ã‚„é–¢ç¯€ãƒˆãƒ«ã‚¯ã�ªã�©ã�®é–¢ç¯€ç©ºé–“ã�Œç”¨ã�„ã‚‰ã‚Œã‚‹\cite{margolis2024rapid, ha2025learning}.
ä¸€èˆ¬ã�«,é–¢ç¯€ç©ºé–“ã�‹ã‚‰ã‚¿ã‚¹ã‚¯ç©ºé–“ã�¸ã�®å†™åƒ�ã�¯,æŽ¥è§¦çŠ¶æ…‹ã‚„é–¢ç¯€é…�ç½®ã�¨ã�„ã�£ã�Ÿé›¢æ•£çš„ã�‹ã�¤é�žç·šå½¢ã�ªè¦�å› ã�«å¼·ã��å½±éŸ¿ã�•ã‚Œã‚‹ã�Ÿã‚�,ã��ã�®å¯¾å¿œé–¢ä¿‚ã�¯è¤‡é›‘ã�«ã�ªã‚‹.
ã��ã�®çµ�æžœ,å­¦ç¿’é€Ÿåº¦ã�®ä½Žä¸‹ã‚„å±€æ‰€è§£ã�¸ã�®å�Žæ�Ÿã‚’æ‹›ã��ã‚„ã�™ã�„.
ã�“ã�®å•�é¡Œã�«å¯¾ã�—,å¾“æ�¥ã�®ç ”ç©¶ã�§ã�¯ç’°å¢ƒè¨­è¨ˆã‚„å ±é…¬è¨­è¨ˆã�®å·¥å¤«ã�«ã‚ˆã�£ã�¦å¯¾å‡¦ã�Œå›³ã‚‰ã‚Œã�¦ã��ã�Ÿ\cite{vogel2024robust_ladder,bellegarda2024robust,yoneda2025kleiyn}.

ã�—ã�‹ã�—,ã‚ˆã‚Šç›´æŽ¥çš„ã�ªè§£æ±ºç­–ã�¨ã�—ã�¦,ã‚¢ã‚¯ã‚·ãƒ§ãƒ³ç©ºé–“ã‚’ã‚¿ã‚¹ã‚¯ç©ºé–“ã�«è¿‘ã�¥ã�‘ã‚‹ã‚¢ãƒ—ãƒ­ãƒ¼ãƒ�ã�Œæœ‰åŠ¹ã�§ã�‚ã‚‹.
ä¾‹ã�ˆã�°ãƒ­ãƒœãƒƒãƒˆã‚¢ãƒ¼ãƒ ã�®åˆ¶å¾¡ã�§ã�¯,ã‚¨ãƒ³ãƒ‰ã‚¨ãƒ•ã‚§ã‚¯ã‚¿ä½�ç½®ã‚’ã‚¢ã‚¯ã‚·ãƒ§ãƒ³ã�¨ã�™ã‚‹ã�“ã�¨ã�§æŽ¢ç´¢åŠ¹çŽ‡ã�Œå�‘ä¸Šã�™ã‚‹ã�“ã�¨ã�Œå ±å‘Šã�•ã‚Œã�¦ã�„ã‚‹\cite{martin2019variable, jiang2024learning, aljalbout2024role}.
è„šãƒ­ãƒœãƒƒãƒˆã�«ã�Šã�„ã�¦ã‚‚,è¶³å…ˆç©ºé–“ã‚’ã‚¢ã‚¯ã‚·ãƒ§ãƒ³ã�¨ã�™ã‚‹æ‰‹æ³•\cite{duan2021learning}ã‚„,è¶³å…ˆä½�ç½®ãƒ»æ­©è¡Œä½�ç›¸ã‚’ä¸Šä½�ã‚¢ã‚¯ã‚·ãƒ§ãƒ³ã�¨ã�—ã�¦ãƒ¢ãƒ‡ãƒ«ãƒ™ãƒ¼ã‚¹åˆ¶å¾¡ã‚’çµ„ã�¿å�ˆã‚�ã�›ã‚‹éšŽå±¤åž‹æ‰‹æ³•\cite{bellegarda2022cpg, kang2025dynamic},å‘¨æœŸè»Œé�“ã�ªã�©ã�®äº‹å‰�æ§‹é€ ã‚’å°Žå…¥ã�™ã‚‹æ‰‹æ³•\cite{miki2022anymal}ã�Œæ��æ¡ˆã�•ã‚Œã�¦ã�„ã‚‹.
ã�“ã‚Œã‚‰ã�®æ‰‹æ³•ã�¯å­¦ç¿’åŠ¹çŽ‡ã�®æ”¹å–„ã�«å¯„ä¸Žã�™ã‚‹ã‚‚ã�®ã�®,è¶³å…ˆç©ºé–“ã�¨ãƒ­ãƒœãƒƒãƒˆæœ¬ä½“ã�®ä½�ç½®ãƒ»å§¿å‹¢ç©ºé–“ã�¨ã�®é–“ã�«ã�¯ä¾�ç„¶ã�¨ã�—ã�¦é�žç·šå½¢æ€§ã�Œå­˜åœ¨ã�™ã‚‹ã�Ÿã‚�,æŽ¢ç´¢æ€§èƒ½ã�®å�‘ä¸Šã�«ã�¯é™�ç•Œã�Œã�‚ã�£ã�Ÿ.
ã�¾ã�Ÿ,å®Ÿæ©Ÿé�©ç”¨ã‚’å‰�æ��ã�¨ã�™ã‚‹è¨­è¨ˆåˆ¶ç´„ã�«ã‚ˆã‚Š,ã‚¢ã‚¯ã‚·ãƒ§ãƒ³è¡¨ç�¾ã�®è‡ªç”±åº¦ã�Œåˆ¶é™�ã�•ã‚Œã‚‹å ´å�ˆã‚‚å¤šã�„.

\begin{figure}[tbp]
    \centering
        \includegraphics[width=0.95\columnwidth]{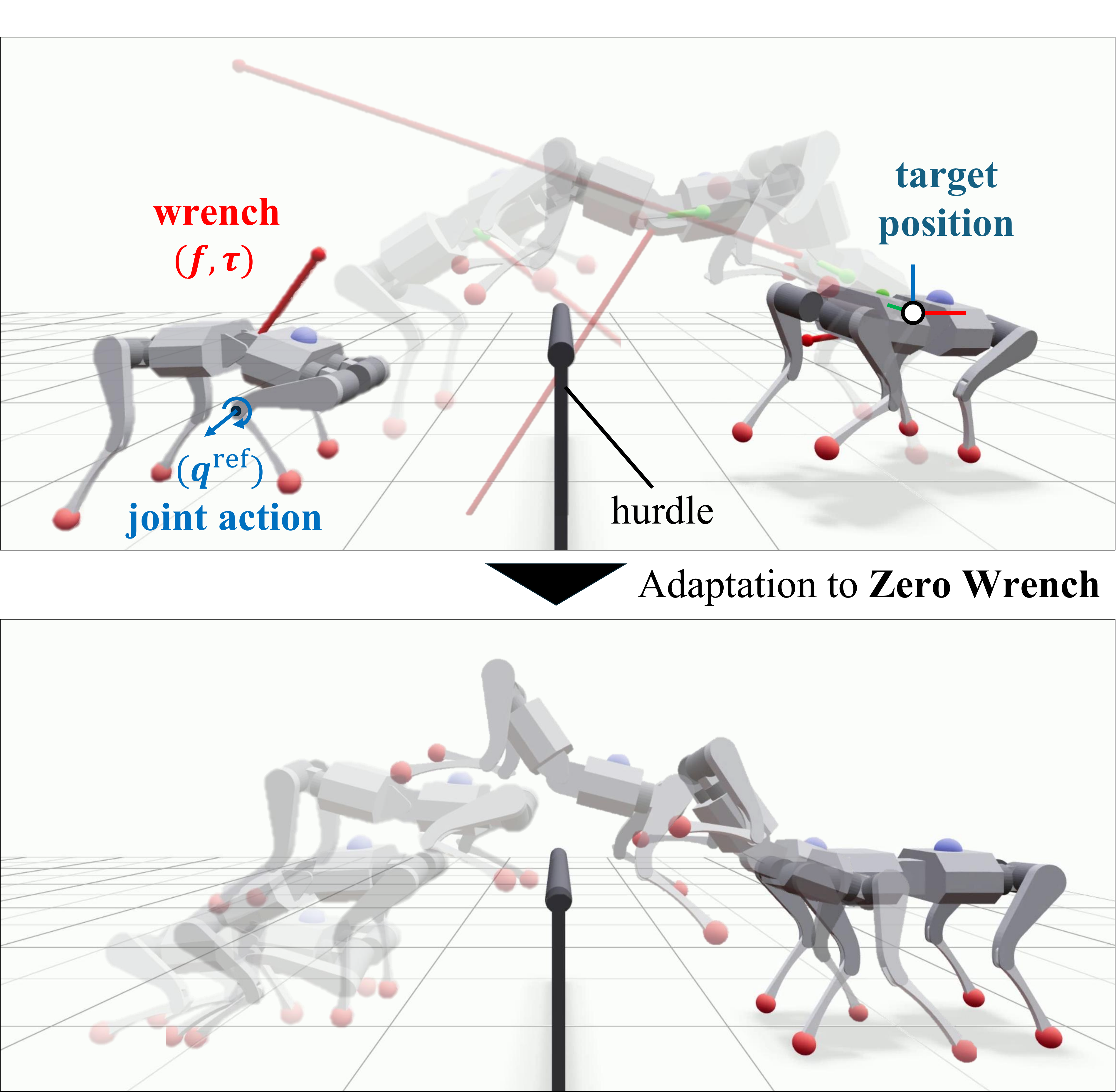}
\caption{Wrench-Augmented Reinforcement Learning (WARL) ã�®æ¦‚è¦�. ãƒ¬ãƒ³ãƒ�ã‚’ã‚¢ã‚¯ã‚·ãƒ§ãƒ³ç©ºé–“ã�«å°Žå…¥ã�™ã‚‹ã�“ã�¨ã�§æŽ¢ç´¢ã‚’ä¿ƒé€²ã�—,ã‚«ãƒªã‚­ãƒ¥ãƒ©ãƒ å­¦ç¿’ã�«ã‚ˆã‚Šæœ€çµ‚çš„ã�«ã�¯ãƒ¬ãƒ³ãƒ�ã‚’ç”¨ã�„ã�ªã�„æ–¹ç­–ã‚’ç�²å¾—ã�™ã‚‹.}
\label{fig:warl_overview}
\vspace{-3ex}
\end{figure}

è„šãƒ­ãƒœãƒƒãƒˆã�®å‹•çš„é�‹å‹•ã�«ã�Šã�„ã�¦æœ¬è³ªçš„ã�ªç‰¹å¾´é‡�ã�®ä¸€ã�¤ã�¯,ãƒ­ãƒœãƒƒãƒˆã�®ä½�ç½®ã�Šã‚ˆã�³å§¿å‹¢ã�®å¤‰åŒ–ã�§ã�‚ã‚‹.
ã�—ã�Ÿã�Œã�£ã�¦,ã�“ã‚Œã‚‰ã�«ç›´æŽ¥ä½œç”¨ã�™ã‚‹ãƒ¬ãƒ³ãƒ�$\bm{w} = (\bm{f}, \bm{\tau}) \in \mathbb{R}^6$ã‚’ã‚¢ã‚¯ã‚·ãƒ§ãƒ³ç©ºé–“ã�«çµ„ã�¿è¾¼ã‚€ã�“ã�¨ã�§,ã‚¿ã‚¹ã‚¯ç©ºé–“ã�¨ã�®å¯¾å¿œé–¢ä¿‚ã‚’å�˜ç´”åŒ–ã�—,æŽ¢ç´¢ã‚’ä¿ƒé€²ã�§ã��ã‚‹å�¯èƒ½æ€§ã�Œã�‚ã‚‹.
ãƒ¬ãƒ³ãƒ�ã‚’é�‹å‹•å­¦ç¿’ã�«åˆ©ç”¨ã�™ã‚‹ç ”ç©¶ã�¨ã�—ã�¦ã�¯,åŠ›å­¦ãƒ¢ãƒ‡ãƒ«ã�«åŸºã�¥ã�„ã�¦å�‚ç…§è»Œé�“ã�¸ã�®è¿½å¾“ã‚’è£œåŠ©ã�™ã‚‹ãƒ¬ãƒ³ãƒ�ã‚’è¿½åŠ ã�™ã‚‹æ‰‹æ³•(ZEST)\cite{sleiman2026zest}ã‚„,ãƒ¬ãƒ³ãƒ�ã‚’å‡ºåŠ›ã�™ã‚‹ãƒ�ãƒªã‚·ãƒ¼ã‚’åˆ©ç”¨ã�—ã�¦å­¦ç¿’ã‚’åŠ é€Ÿã�™ã‚‹æ‰‹æ³•(A2CF)\cite{cao2025learning}ã�ªã�©ã�Œæ��æ¡ˆã�•ã‚Œã�¦ã�„ã‚‹.
ã�—ã�‹ã�—ZESTã�§ã�¯ãƒ¬ãƒ³ãƒ�ã�®è¨ˆç®—ã�¯ãƒ¢ãƒ‡ãƒ«ãƒ™ãƒ¼ã‚¹ã�§ã�‚ã‚Š,æŽ¢ç´¢ã�«ã�¯ç”¨ã�„ã‚‰ã‚Œã�¦ã�„ã�ªã�„.
ã�¾ã�ŸA2CFã�§ã�¯ãƒ¬ãƒ³ãƒ�ã�¯è£œåŠ©ã�¨ã�—ã�¦ã�®æ´»ç”¨ã�«é‡�ã��ã�Œç½®ã�‹ã‚Œã�¦ã�Šã‚Š,æŽ¢ç´¢ã‚’ä¿ƒé€²ã�™ã‚‹ã‚¢ã‚¯ã‚·ãƒ§ãƒ³ã�¨ã�—ã�¦ã�®ã‚·ã‚¹ãƒ†ãƒ è¨­è¨ˆã�¯è¡Œã‚�ã‚Œã�¦ã�„ã�ªã�„.
ã�•ã‚‰ã�«,ãƒ¬ãƒ³ãƒ�ã‚’ç”¨ã�„ã�ªã��ã�¦ã‚‚å­¦ç¿’å�¯èƒ½ã�ªã‚¿ã‚¹ã‚¯ã�§ã�®æ¯”è¼ƒã�«ç•™ã�¾ã�£ã�¦ã�„ã‚‹ã�Ÿã‚�,ãƒ¬ãƒ³ãƒ�ã‚’æŽ¢ç´¢èª˜å°Žã�®ä¸»è¦�ã�ªã‚¢ã‚¯ã‚·ãƒ§ãƒ³ã�¨ã�—ã�¦å°Žå…¥ã�™ã‚‹ã�“ã�¨ã�®æœ‰åŠ¹æ€§ã�¯,ã�“ã‚Œã�¾ã�§å��åˆ†ã�«æ¤œè¨¼ã�•ã‚Œã�¦ã�“ã�ªã�‹ã�£ã�Ÿ.

ã��ã�“ã�§æœ¬ç ”ç©¶ã�§ã�¯,ãƒ¬ãƒ³ãƒ�ã‚’æŽ¢ç´¢èª˜å°Žã�®ã�Ÿã‚�ã�®ä¸»è¦�ã�ªã‚¢ã‚¯ã‚·ãƒ§ãƒ³ã�¨ã�—ã�¦å°Žå…¥ã�—,ã‚«ãƒªã‚­ãƒ¥ãƒ©ãƒ å­¦ç¿’ã�«ã‚ˆã�£ã�¦æœ€çµ‚çš„ã�«ãƒ¬ãƒ³ãƒ�ã�¸ã�®ä¾�å­˜ã‚’ã�ªã��ã�™æ‰‹æ³•ã‚’æ��æ¡ˆã�™ã‚‹.
å­¦ç¿’åˆ�æœŸã�¯ãƒ¬ãƒ³ãƒ�ä¸»ä½“ã�§é�‹å‹•ã‚’ç�²å¾—ã�—,å­¦ç¿’ã�®é€²è¡Œã�«ä¼´ã�„ã��ã�®å¯„ä¸Žã‚’æ®µéšŽçš„ã�«ä½Žæ¸›ã�•ã�›ã‚‹.
å…·ä½“çš„ã�«ã�¯,ãƒ�ãƒªã‚·ãƒ¼ã�Œå‡ºåŠ›ã�—ã�Ÿãƒ¬ãƒ³ãƒ�ã‚’ãƒ­ãƒœãƒƒãƒˆã�®èƒ´ä½“ã�«ç›´æŽ¥ä½œç”¨ã�•ã�›ã‚‹ã�“ã�¨ã�§,ç©ºé–“ç§»å‹•ã‚¿ã‚¹ã‚¯ã�«ã�Šã�„ã�¦é«˜ã�„æŽ¢ç´¢æ€§èƒ½ã‚’å®Ÿç�¾ã�—,æœ€çµ‚çš„ã�«ã�¯ãƒ¬ãƒ³ãƒ�ã‚’ç”¨ã�„ã�šã�«å�Œç­‰ã�®é�‹å‹•ã‚’ç”Ÿæˆ�å�¯èƒ½ã�ªé–¢ç¯€åˆ¶å¾¡æ–¹ç­–ã�®ç�²å¾—ã‚’ç›®æŒ‡ã�™.

æœ¬ç ”ç©¶ã�®ä¸»ã�ªè²¢çŒ®ã�¯ä»¥ä¸‹ã�®é€šã‚Šã�§ã�‚ã‚‹.
\begin{enumerate}
\item è„šãƒ­ãƒœãƒƒãƒˆã�®é�‹å‹•å­¦ç¿’ã�«ã�Šã�„ã�¦,ã‚¢ã‚¯ã‚·ãƒ§ãƒ³ç©ºé–“ã�«ãƒ¬ãƒ³ãƒ�ã‚’å°Žå…¥ã�™ã‚‹ã�“ã�¨ã�§æŽ¢ç´¢ã‚’åŠ¹çŽ‡åŒ–ã�—,ã‚¿ã‚¹ã‚¯å›ºæœ‰ã�®ç’°å¢ƒè¨­è¨ˆã‚„å ±é…¬è¨­è¨ˆã�«å¤§ã��ã��ä¾�å­˜ã�—ã�ªã�„,æ±Žç”¨çš„ã�ªå­¦ç¿’ã�Œå�¯èƒ½ã�«ã�ªã‚‹ã�“ã�¨ã‚’ç¤ºã�—ã�Ÿ.
\item æŽ¢ç´¢èª˜å°Žã�¨ã�—ã�¦å°Žå…¥ã�—ã�Ÿãƒ¬ãƒ³ãƒ�ã‚’æ®µéšŽçš„ã�«é™¤åŽ»ã�—,å®Ÿæ©Ÿé�©ç”¨ã‚’å�¯èƒ½ã�«ã�™ã‚‹æ–¹ç­–ã‚’ç�²å¾—ã�™ã‚‹ã�Ÿã‚�ã�®ã‚«ãƒªã‚­ãƒ¥ãƒ©ãƒ å­¦ç¿’æ‰‹æ³•ã‚’æ��æ¡ˆã�—ã�Ÿ.
\item æœ¬æ‰‹æ³•ã�«ã‚ˆã‚‹ç©ºé–“æŽ¢ç´¢èƒ½åŠ›ã�®å�‘ä¸Šã�¨å¼•ã��æ�›ã�ˆã�«,è¶³å ´ã�®ã‚ˆã�†ã�ªç’°å¢ƒã‚’åˆ©ç”¨ã�—ã�Ÿå™¨ç”¨ã�ªå‹•ä½œã�®ç�²å¾—ã�Œå›°é›£ã�«ã�ªã‚‹ã�¨ã�„ã�†ãƒˆãƒ¬ãƒ¼ãƒ‰ã‚ªãƒ•ã�Œå­˜åœ¨ã�™ã‚‹ã�“ã�¨ã‚‚æ˜Žã‚‰ã�‹ã�«ã�—ã�Ÿ.ã�“ã�®ãƒˆãƒ¬ãƒ¼ãƒ‰ã‚ªãƒ•ã�®è§£æ¶ˆã�Œä»Šå¾Œã�®èª²é¡Œã�§ã�‚ã‚‹.
\end{enumerate}
}%

\section{Background}\label{sec:background}

\subsection{Difficulty of Exploration in Reinforcement Learning for Legged Robots}
Reinforcement learning is a framework for optimizing a policy through interaction with an environment, and is typically formulated as a Markov Decision Process (MDP).
An MDP consists of a state $s$, an action $a$, a reward $r$, and a transition distribution, and the agent learns a policy $\pi(a|s)$ that maximizes the expected cumulative reward.
In motor learning for legged robots, the state includes the robot's posture, velocity, and contact information, and the action is output as joint torques or joint angles.

Proximal Policy Optimization (PPO) \cite{schulman2017ppo} is widely used as an algorithm for reinforcement learning in legged robots.
PPO is an algorithm that ensures learning stability by constraining the policy update amount, and it is also easy to combine with curriculum learning.
Therefore, this study also adopts PPO as the baseline algorithm.
In PPO, which assumes a stochastic policy, exploration is performed by sampling actions based on a Gaussian distribution:

\begin{equation}
a_t \sim \pi_\theta(a_t|s_t) = \mathcal{N}(\mu_\theta(s_t), \Sigma_\theta(s_t))
\end{equation}

Since this randomness-based exploration is performed in the action space, its design greatly affects exploration efficiency.
Conventional joint-space actions are typically of the form where the target joint angles are output by a scale transformation as follows, and then converted to joint torques by PD control.
\begin{equation}
\bm{q}^{\text{ref}} = \bm{q}^{\text{range}} \odot \text{clip}(\lambda^{\text{joint}} \bm{a}^{\text{joint}}, -1, 1) + \bm{q}^{\text{offset}} \label{eq:eq_joint_scale}
\end{equation}
Here, $\bm{a}$ is the action output by the policy, which is converted to the target joint angle $\bm{q}^{\text{ref}}$ using a scaling factor $\lambda^{\text{joint}}$, an action range $\bm{q}^{\text{range}}$, and an offset $\bm{q}^{\text{offset}}$.

\subsection{Difference in Spatial Exploration Capability due to Action Space Design}
\begin{figure}[tbp]
\centering
\includegraphics[width=0.95\columnwidth]{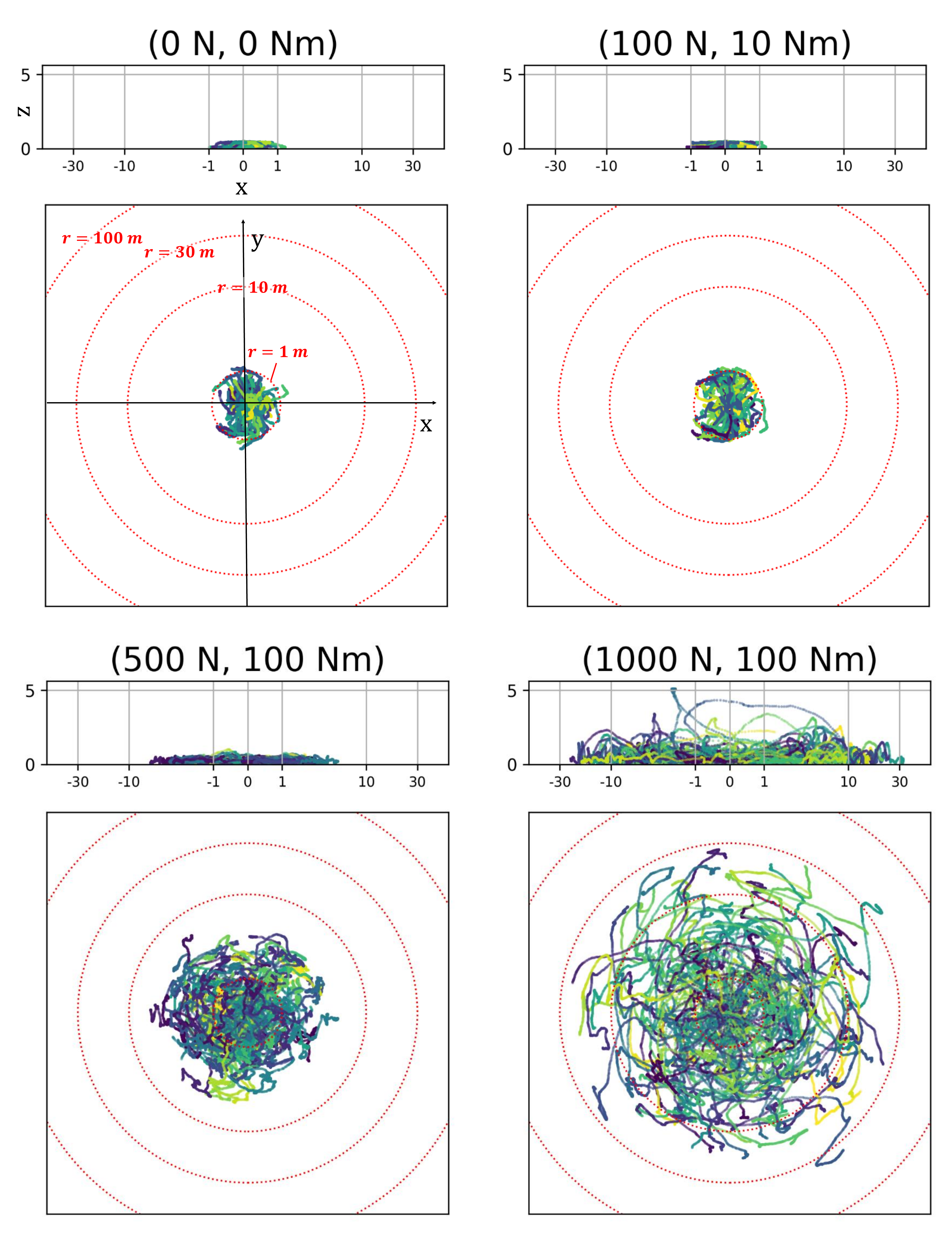}
\caption{Comparison of exploration behavior with different wrench scales. The larger the wrench scale, the more direct exploration in the robot's position space becomes possible, and the spatial exploration capability is dramatically improved.}
\label{fig:exploration_compare}
\vspace{-3ex}
\end{figure}

We investigate the effect of action space design on spatial exploration capability by comparing a conventional joint-space action with an action space that includes a wrench.
Using the quadruped robot KLEIYN \cite{yoneda2025kleiyn} in Isaac Gym, we compare the travel distance when Gaussian noise is directly injected into the action outputs before scaling.
The joint angle action is scaled to a joint angle in the form of \equref{eq:eq_joint_scale} and then converted to a joint torque by PD control.
The wrench action is calculated directly by scale transformation and clipping as follows.
\begin{eqnarray}
\bm{w} &=& \bm{w}^{\text{max}}\cdot \mathrm{clip} (\lambda^{\text{wrench}} \bm{a}^{\text{wrench}},-1,1) \label{eq:eq_wrench_action_tmp} \\
\bm{w}^\text{max} &=& (\bm{f}^{\text{max}} \in \mathbb{R}^3, \boldsymbol{\tau}^{\text{max}} \in \mathbb{R}^3) \in \mathbb{R}^6 \label{eq:eq_wrench_scale}
\end{eqnarray}
Here, $\bm{a}^{\text{wrench}}$ is the wrench action, which is converted to the final wrench $\bm{w}$ using the scaling factor $\lambda^{\text{wrench}}$ and the maximum wrench value $\bm{w}^{\text{max}}$.
Here, we set four types of wrench scales $(f^{\text{max}} [N], \tau^{\text{max}}[Nm]) = \{(0, 0), (100, 10), (500, 100), (1000, 100)\}$ and compare the exploration range under each condition.
The action scales were set to $\lambda^\text{joint}=0.2, \lambda^{\text{wrench}}=0.5$, and the joint angle and wrench actions $(\bm{a}^\text{joint}, \bm{a}^\text{wrench})$ were independently sampled according to a standard normal distribution $\mathcal{N}(\bm{0}, \bm{I})$.
Figure \ref{fig:exploration_compare} shows the trajectories when this random action was applied to 100 robots for 10 seconds.

In the case of only joint angles with the wrench disabled ($0 \ N, 0 \ Nm$) and when the wrench scale was small ($100 \ N, 10 \ Nm$), the robots could only move within a range of about 1m in 10 seconds.
In contrast, with a large wrench scale ($500 \ N, 100 \ Nm$), trajectories of more than 10m in 10 seconds were observed, and with an even larger scale ($1000 \ N, 100 \ Nm$), vertical displacement also increased substantially.

These results indicate that by including a wrench in the action space, it becomes possible to explore directly in the robot's position space with only Gaussian noise, and the spatial exploration capability is dramatically improved.

\subsection{Gradual Attenuation of Wrench using Curriculum Learning}

Although introducing a wrench improves exploration efficiency, it is not possible to directly apply a wrench to the robot's body in reality.
Therefore, previous studies \cite{cao2025learning, sleiman2026zest} have proposed curriculum learning that gradually attenuates the wrench output based on indicators such as success rate and the magnitude of the auxiliary wrench.

In this study, we follow the framework of previous studies that gradually attenuate the wrench and propose a curriculum learning method that uses only the success rate as an indicator.

\section{Method}\label{sec:method}

\subsection{Overview of WARL}

\begin{figure}[tbp]
\centering
\includegraphics[width=0.95\columnwidth]{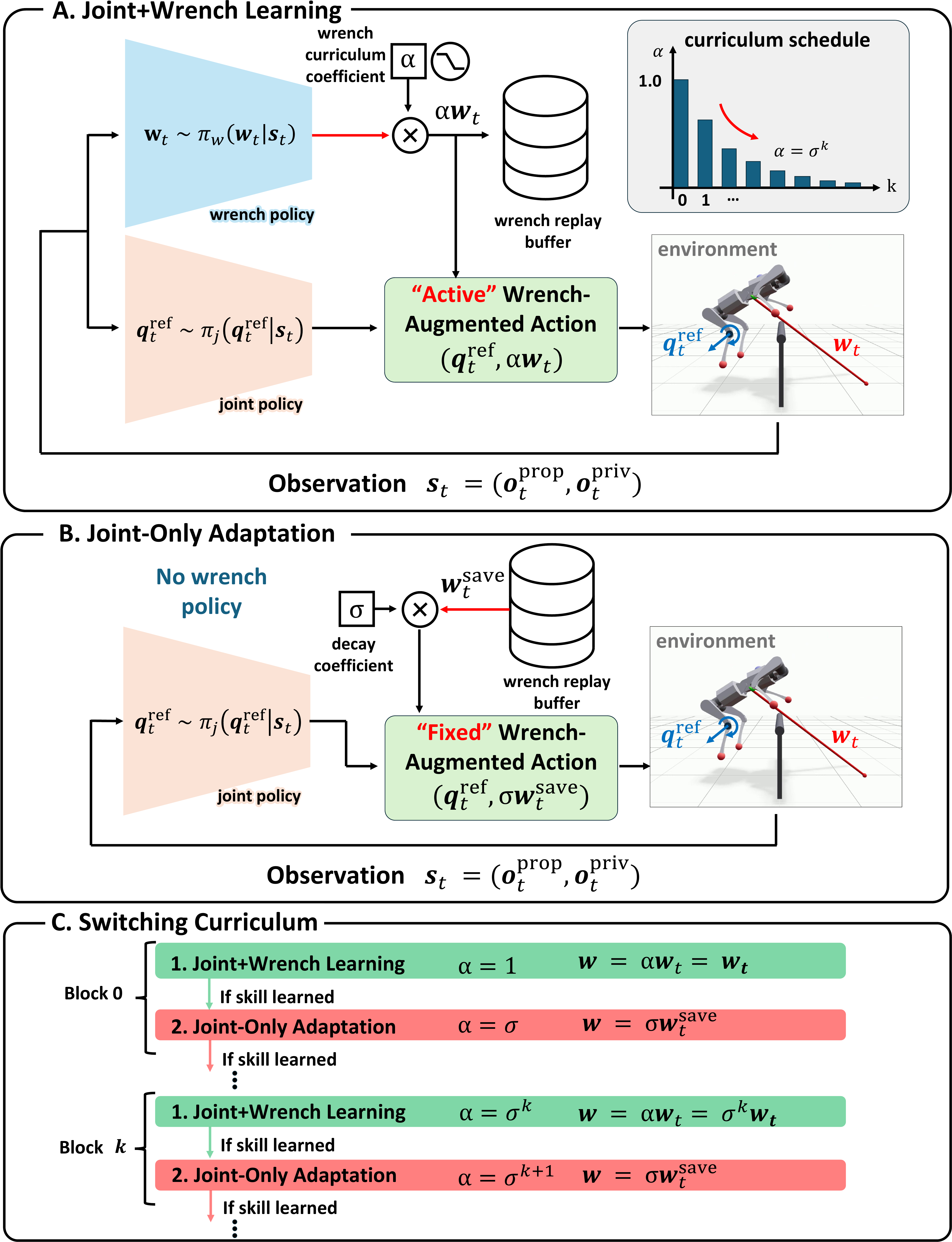}
\caption{Learning flow of WARL. Joint+Wrench learning promotes exploration and stores successful wrench sequences. Joint-only adaptation reduces wrench dependence using a decayed coefficient $\alpha$.}
\label{fig:warl_detail}
\vspace{-3ex}
\end{figure}

We propose Wrench-Augmented Reinforcement Learning (WARL), a framework that exploits wrench-based exploration while ultimately acquiring a joint-only control policy.
As illustrated in Fig.~\ref{fig:warl_detail}, WARL alternates between two phases: (i) a Joint+Wrench learning phase, where both a wrench policy $\pi_w$ and a joint policy $\pi_j$ are optimized, and (ii) a Joint-only adaptation phase, where the motion is reproduced by $\pi_j$ under attenuated previously saved wrenches.
This Switching Curriculum enables efficient spatial exploration in early learning and gradually removes the dependency on the wrench through a decaying curriculum coefficient $\alpha$.
To make the RL problem tractable, we adopt a Teacher-Student framework \cite{miki2022anymal}, where privileged observations are used during training and removed through a student distillation process.

\begin{algorithm}[t]
\caption{Wrench-Augmented Reinforcement Learning (WARL)}
\label{alg:warl}
\begin{algorithmic}[1]
\State Initialize $\pi_w, \pi_j$, $\alpha=1$
\While{$\alpha>\epsilon$}

\State \textbf{Joint+Wrench Learning}
\While{$\zeta<\gamma$ or $|\Delta\zeta|>\delta$}
\State $\bm{q}_t^{\mathrm{ref}}=\pi_j(\bm{s}_t)$
\State $\bm{w}_t=\pi_w(\bm{s}_t)$
\State $\bm{s}_{t+1},r_t=\mathrm{Step}(\bm{q}_t^{\mathrm{ref}},\alpha\bm{w}_t)$
\State Update $\pi_j,\pi_w$
\State Store wrench sequences from successful episodes
\EndWhile

\State Freeze $\pi_w$ and set $\alpha\leftarrow\sigma\alpha$

\State \textbf{Joint-only Adaptation}
\While{$\zeta<\gamma$ or $|\Delta\zeta|>\delta$}
\State $\bm{q}_t^{\mathrm{ref}}=\pi_j(\bm{s}_t)$
\State $\bm{w}_t=\mathcal{W}_{\mathrm{saved}}(t)$
\State $\bm{s}_{t+1},r_t=\mathrm{Step}(\bm{q}_t^{\mathrm{ref}},\sigma\bm{w}_t)$
\State Update $\pi_j$
\EndWhile
\EndWhile
\State \Return $\pi_j$
\end{algorithmic}
\end{algorithm}

\subsection{Action Space Design}

The wrench policy $\pi_w$ outputs a 7-dimensional vector:
\begin{equation}
(\bm{a}^{\text{force}}\in\mathbb{R}^3,\,
\bm{a}^{\text{torque}}\in\mathbb{R}^3,\,
a^{\text{scale}}\in\mathbb{R})=\pi_w(\bm{s}).
\end{equation}

The wrench components are computed as
\begin{eqnarray}
\lambda^{\text{scale}} &=& \mathrm{sigmoid}(a^{\text{scale}}), \\
\bm{f} &=& f^{\text{max}}\cdot
\mathrm{clip}(\lambda^{\text{wrench}}\bm{a}^{\text{force}},-1,1)
\cdot\lambda^{\text{scale}}\cdot\alpha, \\
\boldsymbol{\tau} &=& \tau^{\text{max}}\cdot
\mathrm{clip}(\lambda^{\text{wrench}}\bm{a}^{\text{torque}},-1,1)
\cdot\lambda^{\text{scale}}\cdot\alpha.
\end{eqnarray}

Here, $\lambda^{\text{wrench}}$ is a fixed scaling factor and $\alpha$ is the curriculum coefficient.
The resulting wrench is transformed to the global frame using the torso orientation $R\in \mathrm{SO}(3)$ and applied to the torso link.

\subsection{Switching Curriculum}

Optimizing joint control and wrench-guided exploration simultaneously can cause training instability. To address this, we propose a Switching Curriculum that alternates between the following two phases.

\textbf{Joint+Wrench Phase:}
Both $\pi_w$ and $\pi_j$ are optimized under the scaled wrench $\alpha\bm{w}_t$.
Successful wrench sequences are stored for subsequent adaptation.
This phase primarily enhances spatial exploration.

\textbf{Joint-only Adaptation Phase:}
The joint policy $\pi_j$ is trained using attenuated saved wrenches $\sigma\mathcal{W}_{\mathrm{saved}}$ to encourage reproduction of the motion with reduced wrench assistance.

Each phase terminates when the success rate $\zeta$ exceeds $\gamma$ and its change $|\Delta\zeta|$ falls below $\delta$.
After adaptation, $\alpha$ is updated as $\alpha\leftarrow\sigma\alpha$.
This process continues until $\alpha<\epsilon$, yielding a joint-only policy.

\subsection{Saving Wrench Sequences}

For each successful episode $i$, we define the 6D wrench sequence $\mathcal{W}^i$, its norm $|\mathcal{W}^i|$, and the sum of two sequences $\mathcal{W}^A + \mathcal{W}^B$ as

\begin{eqnarray}
	\mathcal{W}^i=\{\bm{w}_t^i\}_{t=0}^{T_i}=\{(\bm{f}_t^i, \bm{\tau}_t^i)\}_{t=0}^{T_i}. \\
|\mathcal{W}^i|=\sum_{t=0}^{T_i}
\sqrt{||\bm{f}_t^i||^2+\kappa||\boldsymbol{\tau}_t^i||^2}. \\
	\mathcal{W}^A + \mathcal{W}^B = \{\bm{w}_t^A + \bm{w}_t^B\}_{t=0}^{T}.
\end{eqnarray}

$\kappa$ is a weighting factor to balance force and torque magnitudes.
When the buffer accumulates $N^{\text{succ}}$ sequences, the median sequence is selected and the saved wrench is updated:
\begin{equation}
\mathcal{W}_{\mathrm{saved}}^{j}
=\eta\mathcal{W}_{\mathrm{saved}}^{j-1}
+(1-\eta)\mathcal{W}^{i^\ast}.
\end{equation}

This median-based weighted update suppresses extreme cases and stabilizes adaptation.

\subsection{Teacher-Student Learning}

Learning legged locomotion using only proprioceptive inputs can be unstable due to the high-dimensional state space and sparse rewards. To address this, we adopt a teacher-student framework~\cite{miki2022anymal} that stabilizes training by leveraging privileged observations.

\section{Learning Setup}\label{sec:learning_setup}

\subsection{Robot Models}

To verify the versatility of our method, we use two quadruped robot models with different sizes and joint configurations: KLEIYN (total length 750 mm, weight 18 kg, max joint torque 24 Nm, max joint speed 480 rpm) and Unitree Go2 (total length 700 mm, weight 15 kg, max joint torque 24 Nm, max joint speed 230 rpm). We select Isaac Gym~\cite{liang2018isaacgym} as our simulation environment. Specifically, we apply wrenches to the robot's torso link using the \texttt{apply\_rigid\_body\_force\_tensor} method, and control its joints via torque commands using the \texttt{set\_dof\_actuation\_force\_tensor} method.

\subsection{Task Settings}

To verify the effect of exploration guidance by wrench, we set up two types of tasks: goal-reaching tasks and dynamic tasks.

\paragraph{Goal-reaching tasks}

These consist of four tasks: \textsl{HurdleJump}, \textsl{LedgeJump}, \textsl{GapLeap}, and \textsl{IslandTraverse}, all of which require the robot to reach a specified target position $\bm{p}^*$ and target orientation $\bm{q}^*$.
Specifically, in \textsl{HurdleJump}, a 0.5~m obstacle blocks the path to a goal 2~m away. In \textsl{LedgeJump}, the goal is located on a 0.5~m high ledge 2~m away. In \textsl{GapLeap}, the robot must clear a 1~m wide gap to reach a destination 2~m away. In \textsl{IslandTraverse}, the path to a destination 2.4~m away includes a 1.4~m gap containing two 0.2~m square footholds.
The objective of this learning is to acquire motions to overcome these diverse terrains without relying on reward tuning or environmental curricula.
Therefore, a common reward is used for all tasks, and the terrain is also fixed with a fixed height of steps and size of gaps for learning.

To also evaluate the ability to remain stationary, with a probability of 10\% the target position is set to the robot's initial position, effectively requiring the robot to stay in place.

The success rate was calculated based on the criterion that an episode is considered successful if the distance from the target position $|\bm{p}-\bm{p}^*|$ is within 0.2 m and the error from the target orientation $|\bm{\theta}|$ is within $45\tcdegree$.

\paragraph{Dynamic tasks}

We target two dynamic tasks, \textsl{Backflip} and \textsl{Barrel-Roll}, to evaluate the effectiveness of wrench-guided exploration for motions involving drastic attitude changes. The goal is to perform a backflip and a barrel roll, respectively.
Similarly, in these tasks, with a probability of 10\% the target position is set to the robot's start position so that the robot is required to remain at its initial location.

In the Dynamic task, the success rate was calculated based on the criterion that an episode is considered successful if the rotation angle error around the target axis $|\theta|$ is within $22.5\tcdegree$ and the difference between the robot's height $z$ and the target height $z^*$ is within 0.1 m.

\subsection{Reward and Observation design}
We define the observations and rewards for the Goal-reaching tasks and Dynamic tasks below.
The definitions of each variable are shown in \tabref{tab:variable_notation}, and the respective reward functions are shown in \tabref{tab:reward_goal} and \tabref{tab:reward_dynamic}.

\begin{table}[t]
\centering
\caption{Notation used in observation and reward formulation{}}
\label{tab:variable_notation}
\begin{tabular}{l l}
\hline
Symbol & Description \\
\hline
$\psi(x,\lambda)$ & Gaussian shaping kernel : $\exp(-x^2/\lambda^2)$ \\
$w_{\text{cur}}$ & Curriculum weight at iteration $i$ : ${0.01^{0.9999}}^i$ \\
$N_{\text{contact}}$ & Number of penalized contacts \\
$\bm{f}$ & External force component of wrench \\
$\boldsymbol{\tau}$ & External torque component of wrench \\
\hline
$\boldsymbol{\omega}$ & Base angular velocity \\
$\bm{v}$ & Base linear velocity \\
$\bm{g}$ & Projected gravity vector in base frame \\
$g_x, g_y$ & $x,y$ components of $\bm{g}$ \\
\hline
$\bm{p}$ & Robot base position \\
$\bm{p}^*$ & Target position \\
$\Delta\bm{p}=\bm{p}^*-\bm{p}$ & Position error \\
\hline
$\bm{q}$ & Joint configuration \\
$\bm{q}_{\text{def}}$ & Default joint configuration \\
$\dot{\bm{q}}$ & Joint velocity \\
$\bm{e}_{\text{dof}}$ & $\bm{q}-\bm{q}_{\text{def}}$ \\
\hline
$\boldsymbol{\theta}$ & Orientation error (axis-angle form) \\
$\theta$ & Target rotation angle (pitch/roll) \\
$z$ & Base height \\
$z^*$ & Standing height of the robot \\
\hline
$\phi$ & Phase variable : $t^3/(1+t^3)$ \\
$\bm{h}$ & Measured terrain heights $\in \mathbb{R}^{187}$ \\
$\boldsymbol{\omega}_{xz}$ & Angular velocity with zero $y$ component \\
$\boldsymbol{\omega}_{yz}$ & Angular velocity with zero $x$ component \\
\hline
\end{tabular}
\end{table}

\begin{table}[t]
\centering
\caption{Reward terms for goal-reaching tasks{}}
\label{tab:reward_goal}
\begin{tabular}{l l c}
\hline
Term & Formulation & Weight \\
\hline

Target position &
$ \Big( \psi(|\bm{p}-\bm{p}^*|,\,0.5) - \frac{|\bm{p}-\bm{p}^*|^2}{5^2} \Big) $ & $1.0$ \\

Stand pose &
$ \Big(|\boldsymbol{\theta}|/\pi\Big)^2 \; \psi(|\bm{p}-\bm{p}^*|,\,0.1) $ & $-1.0$
\\

Stand DOF &
$ \psi(|\bm{e}_{\text{dof}}|,\,1) \; \psi(|\bm{p}-\bm{p}^*|,\,0.1) \; \psi(|\boldsymbol{\theta}|,\,\pi/2) $ & $1.0$ \\

Angular velocity &
$ |\boldsymbol{\omega}|^2 \,w_{\text{cur}} $ & $-0.03$ \\

Collision &
$ N_{\text{contact}} \,w_{\text{cur}} $ & $-5.0$ \\

No wrench &
$ \big( |\bm{f}|^2 + 100|\boldsymbol{\tau}|^2 \big) \,w_{\text{cur}} $ & $-1\!\times\!10^{-5}$ \\

\hline
\end{tabular}
\end{table}
\begin{table}[htbp]
\centering
\caption{Reward terms for dynamic tasks{}}
\label{tab:reward_dynamic}
\begin{tabular}{l l c}
\hline
Term & Formulation & Weight \\
\hline

Target angle &
$ \psi(2\pi-\theta,\,\pi/8) $ & $1.0$ \\

Stand height &
$ \psi(2\pi-\theta,\,\pi/8) \; \psi(z-z^*,\,0.1) $ & $1.0$ \\

Stand DOF &
$ \psi(|\bm{e}_{\text{dof}}|,\,0.5) \; \psi(2\pi-\theta,\,\pi/8) \; \psi(|\boldsymbol{\theta}|,\,\pi/8) $ & $1.0$ \\

Angular velocity &
$ 
r_{\omega}=
\begin{cases}
|\boldsymbol{\omega}_{xz}|^2 w_{\text{cur}}, & \text{Backflip}\\
|\boldsymbol{\omega}_{yz}|^2 w_{\text{cur}}, & \text{Barrel roll}
\end{cases}
$ & $-0.03$ \\

No tilt &
$ 
r_{\text{tilt}}=
\begin{cases}
|g_y|, & \text{Backflip}\\
|g_x|, & \text{Barrel roll}
\end{cases}
$ & $-0.3$ \\

Collision &
$ N_{\text{contact}} \,w_{\text{cur}} $ & $-5.0$ \\

No wrench &
$ \big( |\bm{f}|^2 + 100|\boldsymbol{\tau}|^2 \big) \,w_{\text{cur}} $ & $-1\!\times\!10^{-5}$ \\

\hline
\end{tabular}
\end{table}

\paragraph{Observation design}

For each task, we define a proprioceptive observation $\bm{o}^{\text{prop}}$ and a privileged observation $\bm{o}^{\text{priv}}$.
The proprioceptive observation includes information that can be obtained from a real robot, such as the robot's posture, velocity, and contact information. The privileged observation includes information that can only be obtained in a simulation environment, such as the robot's position error and terrain information.
Here, $\phi$ is a phase variable introduced for learning stabilization, and is defined as $\phi = t^3/(1+t^3)$, where $t$ is the time step in the episode.

\subparagraph{Goal-reaching task}

\begin{equation*}
\bm{o}^{\text{prop}} = \Big[ \boldsymbol{\omega},\; \bm{g},\; \bm{e}_{\text{dof}},\; \dot{\bm{q}},\; \phi  \Big],\\
\bm{o}^{\text{priv}} = \Big[\Delta\bm{p},\; \boldsymbol{\theta},\; \bm{h},\; \bm{v} \Big]
\end{equation*}

\subparagraph{Dynamic task}

\begin{equation*}
\bm{o}^{\text{prop}} = \Big[ \boldsymbol{\omega},\; \bm{g},\; \bm{e}_{\text{dof}},\; \dot{\bm{q}},\; \phi \Big],\\
\bm{o}^{\text{priv}} = \Big[ \theta,\; (z-z^*),\; \bm{v} \Big] 
\end{equation*}

\subsection{Learning Parameters}
The scaling factor for the wrench action $\lambda^{\text{wrench}}$ was set to 0.5.
Based on the results of the exploration capability experiment in simulation \figref{fig:exploration_compare}, the parameters that characterize the exploration capability of WARL, $(f^{\text{max}}, \tau^{\text{max}})$, were set to $(1000 N, 100 Nm)$.
The parameters of the WARL curriculum were set as follows.
\begin{equation*}
\quad\gamma=0.6, \quad \delta=2e-6, \quad \sigma=0.8, \quad \epsilon=0.01
\end{equation*}
\begin{equation*}
\quad N^{\text{succ}}=100, \quad \eta=0.9, \quad \kappa=100.0
\end{equation*}
Furthermore, the standing height $z^*$ of each robot was set to the base height at the default joint angle $\bm{q}_{\text{def}}$, which was 0.43m for KLEIYN and 0.3m for Unitree Go2.

We used PPO for learning, and trained 4096 robots in parallel for 5000 iterations. Each policy was updated at 50Hz, and one iteration consisted of 48 steps (one episode is 4 seconds). The learning time was about 5 hours on a single RTX 3090 Ti.

\section{Learning Experiments and Results}\label{sec:learning_experiment}

\subsection{Learning Results in Simulation}\label{sec:learning_result}
\begin{figure}[t]
\centering
\includegraphics[width=0.95\columnwidth]{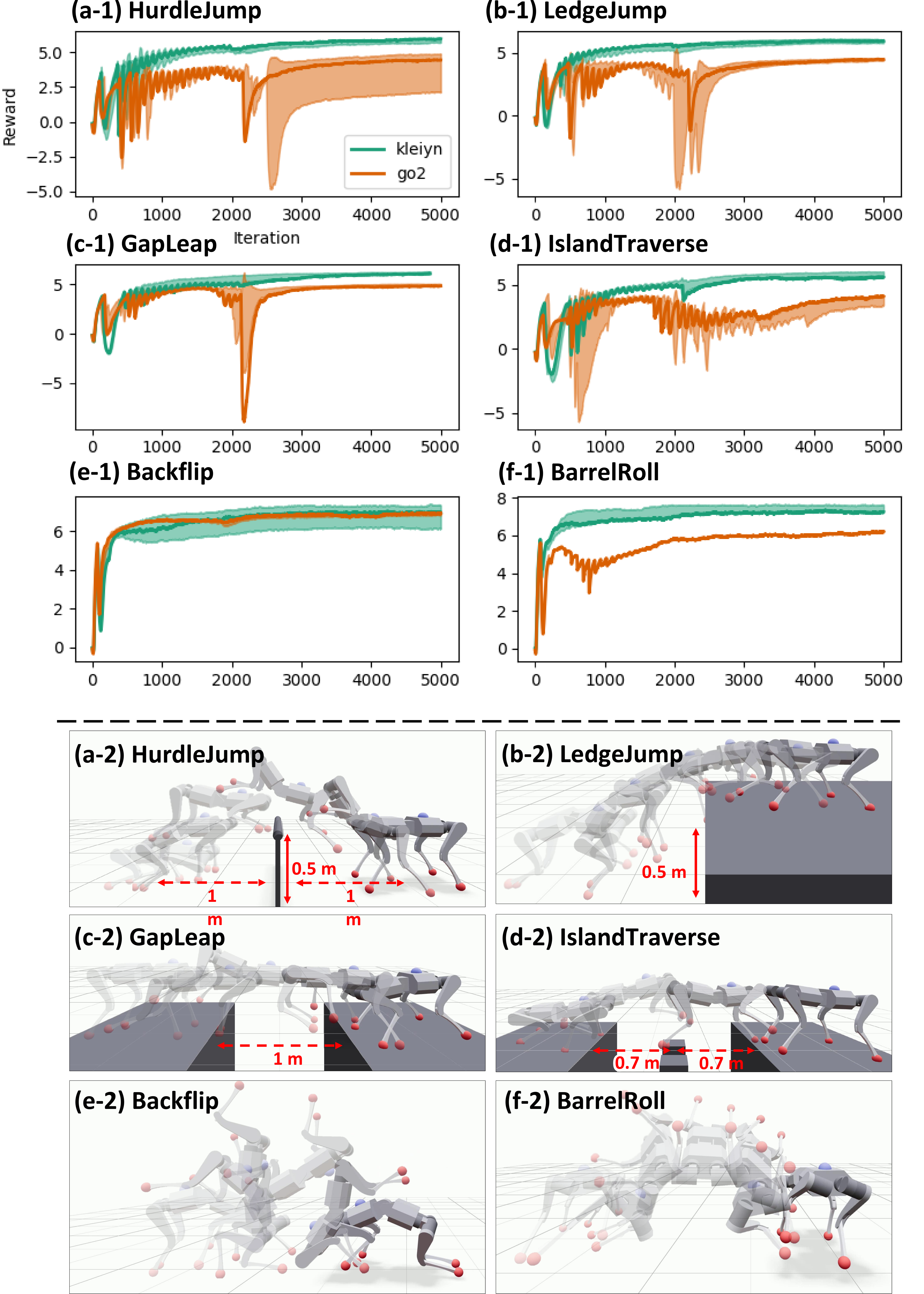}
\caption{[a-f]-1: Transition of rewards when learning six tasks with KLEIYN and Go2, respectively. [a-f]-2: Snapshots of the learned motions in each task.}
\label{fig:kleiyn_go2_result}
\vspace{-2ex}
\end{figure}

Using the proposed method WARL, we trained two robots, KLEIYN and Unitree Go2, on the six tasks defined in \secref{sec:learning_setup} (four Goal-reaching tasks and two Dynamic tasks).
The learning curves and the motions learned by KLEIYN are shown in \figref{fig:kleiyn_go2_result}.
Each graph shows the mean and variance of five trials.
No robot-specific tuning was performed other than changing the standing height $z^*$.

Learning was successful for both robots on a variety of goal-reaching and dynamic motion tasks.
The learning curves show an overall trend of reward improvement, with temporary drops caused by curriculum-driven wrench decay followed by recovery through adaptive learning.
Learning also progressed using the same parameters across different robots.

In the IslandTraverse task, the robot acquired a motion that directly jumped over the 1.4~m gap without using the intermediate footholds.
This likely occurred because wrench-based exploration dominated the learning process, and motions utilizing footholds were not discovered.

\begin{figure}[tbp]
\centering
\includegraphics[width=0.95\columnwidth]{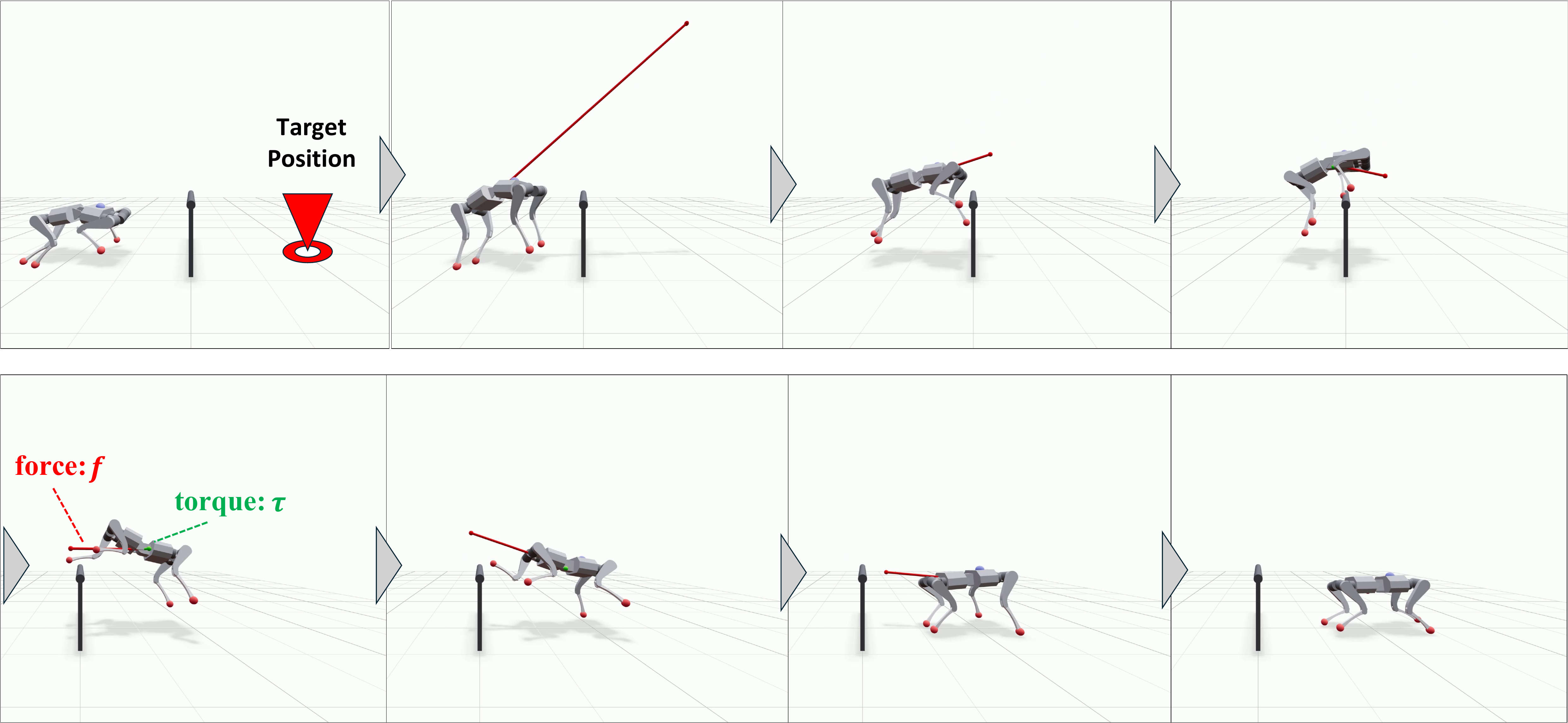}
	\caption{Learned wrench and joint actions in \textsl{HurdleJump} at iteration 500. Wrench assistance is applied to clear the obstacle, while joint control is coordinated to lift the feet.}
\label{fig:learned_wrench}
\vspace{-2ex}
\end{figure}

The learned wrench and joint motion at iteration 500 in the HurdleJump task are shown in \figref{fig:learned_wrench}.
The results indicate that the hurdle-crossing motion is largely generated by the learned wrench, while coordinated leg motions are also acquired to lift the feet and clear the hurdle.

\begin{figure}[tbp]
\centering
\includegraphics[width=0.95\columnwidth]{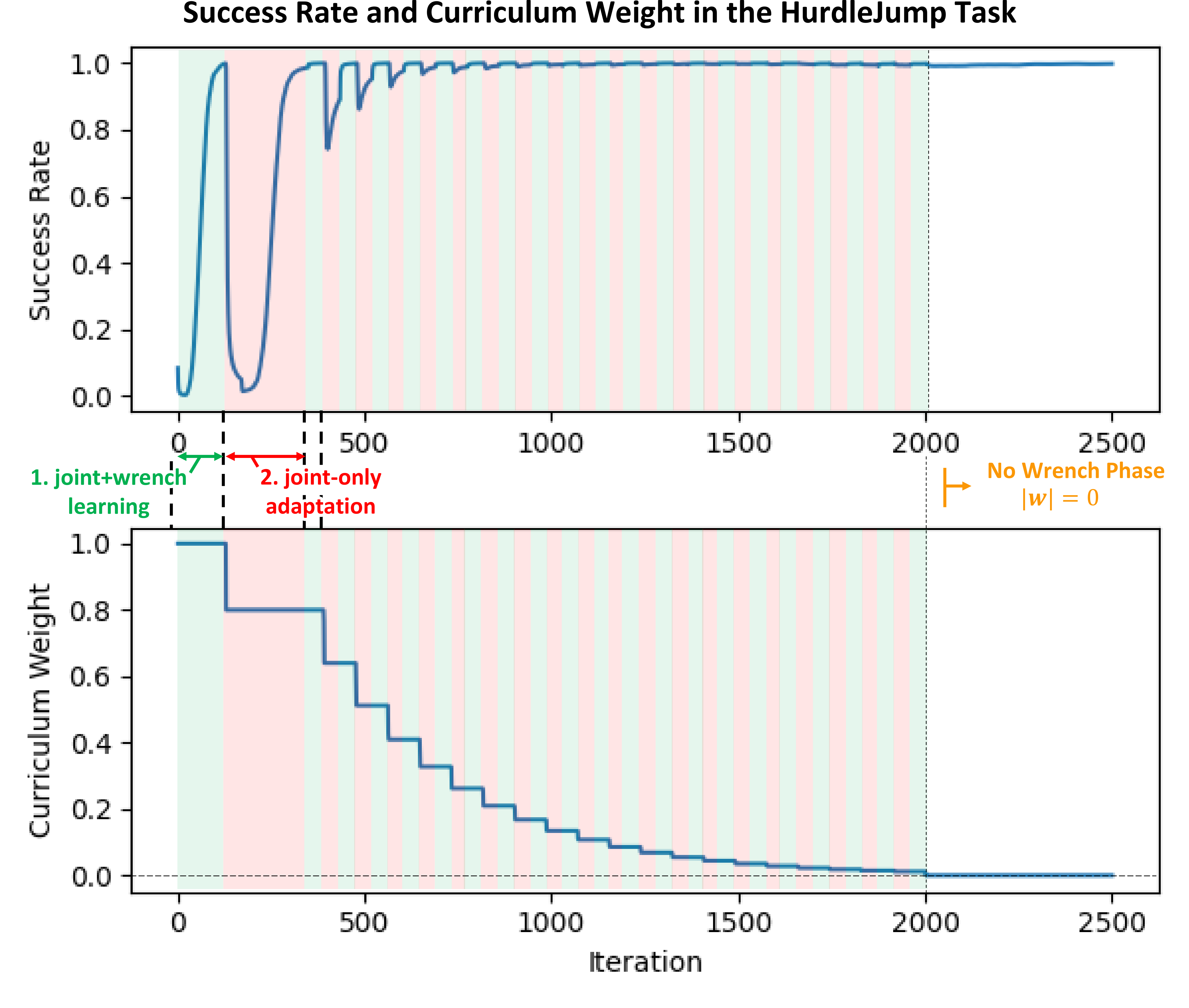}
\vspace{-3ex}
\caption{Transition of success rate and curriculum weight up to 2500 iterations in the HurdleJump task.}
\label{fig:kleiyn_hurdle_warl_result}
\vspace{-3ex}
\end{figure}

To illustrate how the curriculum weight is updated based on the success rate, the transition of the success rate and curriculum weight during learning of the HurdleJump task is shown in \figref{fig:kleiyn_hurdle_warl_result}.

The success rate temporarily decreases when switching from the Joint+Wrench learning phase to the Joint-only adaptive learning phase, but recovers through subsequent adaptive learning and gradually improves.

It takes about 2000 iterations to complete the curriculum, whereas the success rate reaches nearly 1.0 within about 100 iterations in the first Joint+Wrench phase.
This suggests that the target motion itself can be acquired quickly using the wrench.

Most of the training time is spent on adaptive learning to remove wrench dependence and reproduce the motion using only joint control.
This indicates that while the rough structure of the motion can be obtained quickly, improving the efficiency of adaptive learning is important for reducing total training time.

\subsection{Ablation Study}\label{sec:ablation}
To verify the effectiveness of each component of the proposed method, we conduct an ablation study under four conditions: (1) WARL, (2) w/o Switch, (3) w/o Wrench Reward, and (4) Baseline.
(2) w/o Switch: The adaptation phase is omitted, and learning is always performed in the exploration phase.
(3) w/o Wrench Reward: The penalty term for the wrench amount in the exploration phase is removed from the reward function.
(4) Baseline: Training is performed using only PPO and joint control, without including the wrench in the action space. All other training conditions, such as the reward function, are identical to the proposed method. 

\begin{figure}[tbp]
\centering
\includegraphics[width=0.95\columnwidth]{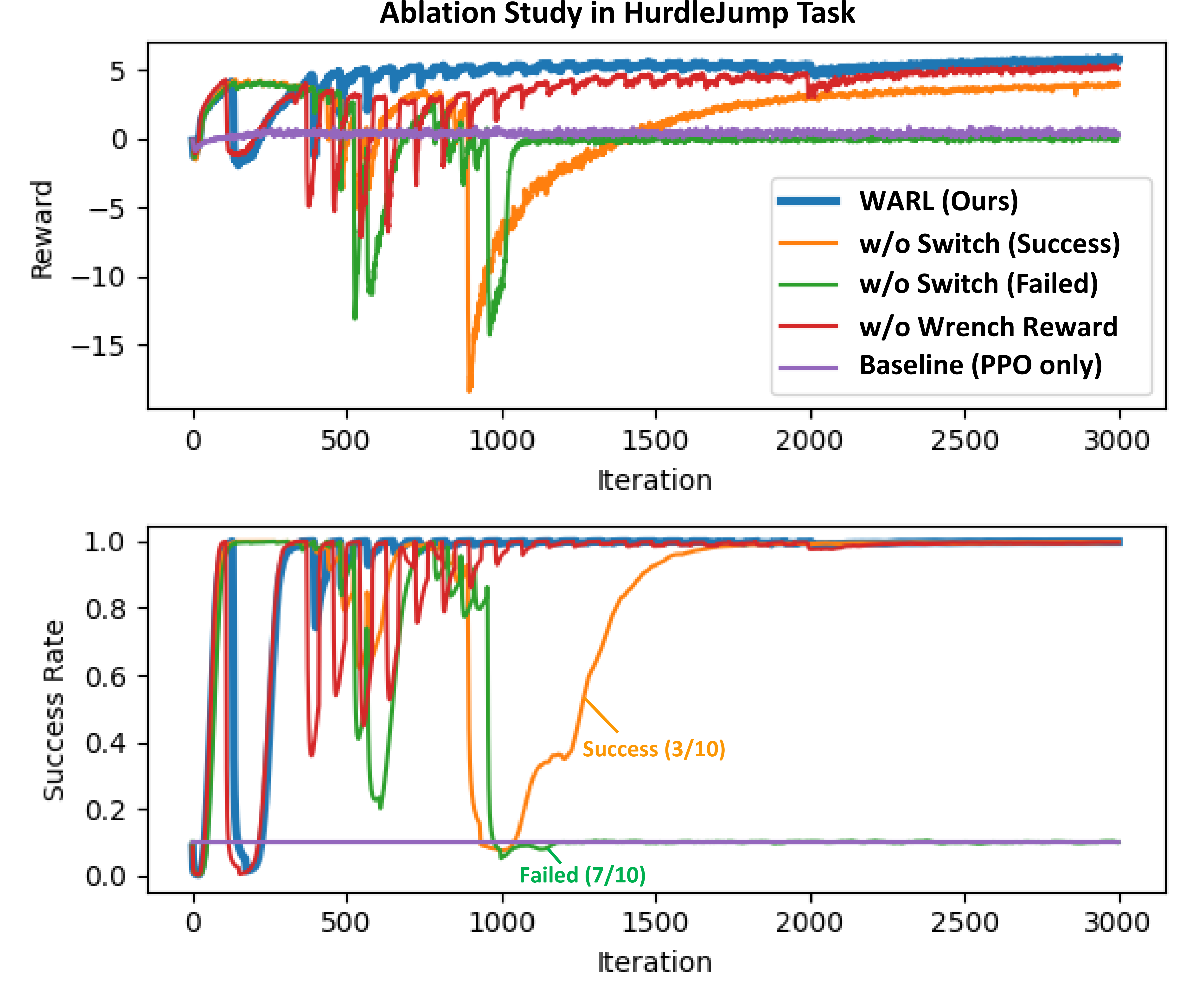}
\vspace{-3ex}
\caption{Results of the ablation study of the proposed method on the HurdleJump task. Learning was performed under four conditions: (1) WARL, (2) w/o Switch, (3) w/o Wrench Reward, and (4) Baseline, and the transition of the success rate is shown. }
\label{fig:ablation}
\vspace{-3ex}
\end{figure}

We learn the HurdleJump task under these conditions and show the transition of the success rate in \figref{fig:ablation}.
Since condition (2) succeeded in only 3 out of 10 trials, both successful and unsuccessful learning curves are shown.

(1) WARL showed the most stable improvement in success rate overall.
In (2) w/o Switch, since the Joint-only adaptation phase was omitted, the learned policy depended on the wrench and the success rate tended to decrease in the final stage of the curriculum.
Learning success after the curriculum was inconsistent, indicating that the Switching Curriculum stabilizes learning.

In (3) w/o Wrench Reward, learning eventually succeeded, but the decrease in success rate at the phase switch was larger than in (1) WARL. Without a penalty on wrench usage, policies with stronger wrench dependence tend to be learned during the Joint+Wrench phase.

In (4) Baseline, since the wrench was not included in the action space, exploration guidance was not obtained and the success rate remained low throughout learning.
This indicates that the task is difficult to explore using only joint control and that wrench-based exploration guidance is highly effective.

These comparisons indicate that the Switching Curriculum plays a central role in removing wrench dependence, while the Wrench Reward helps suppress excessive reliance on the wrench.

\subsection{Execution of Policy for Real Robot in MuJoCo Simulator}

The policy $\pi_j(\bm{a}|\bm{s})$ acquired by WARL includes privileged observation and cannot be directly applied to a real robot. Therefore, we train a policy $\pi_s(\bm{a}|\bm{o}^{\text{prop}})$ that takes only Proprioceptive Observation as input using Teacherâ€“Student learning.

\begin{figure}[tbp]
\centering
\includegraphics[width=0.95\columnwidth]{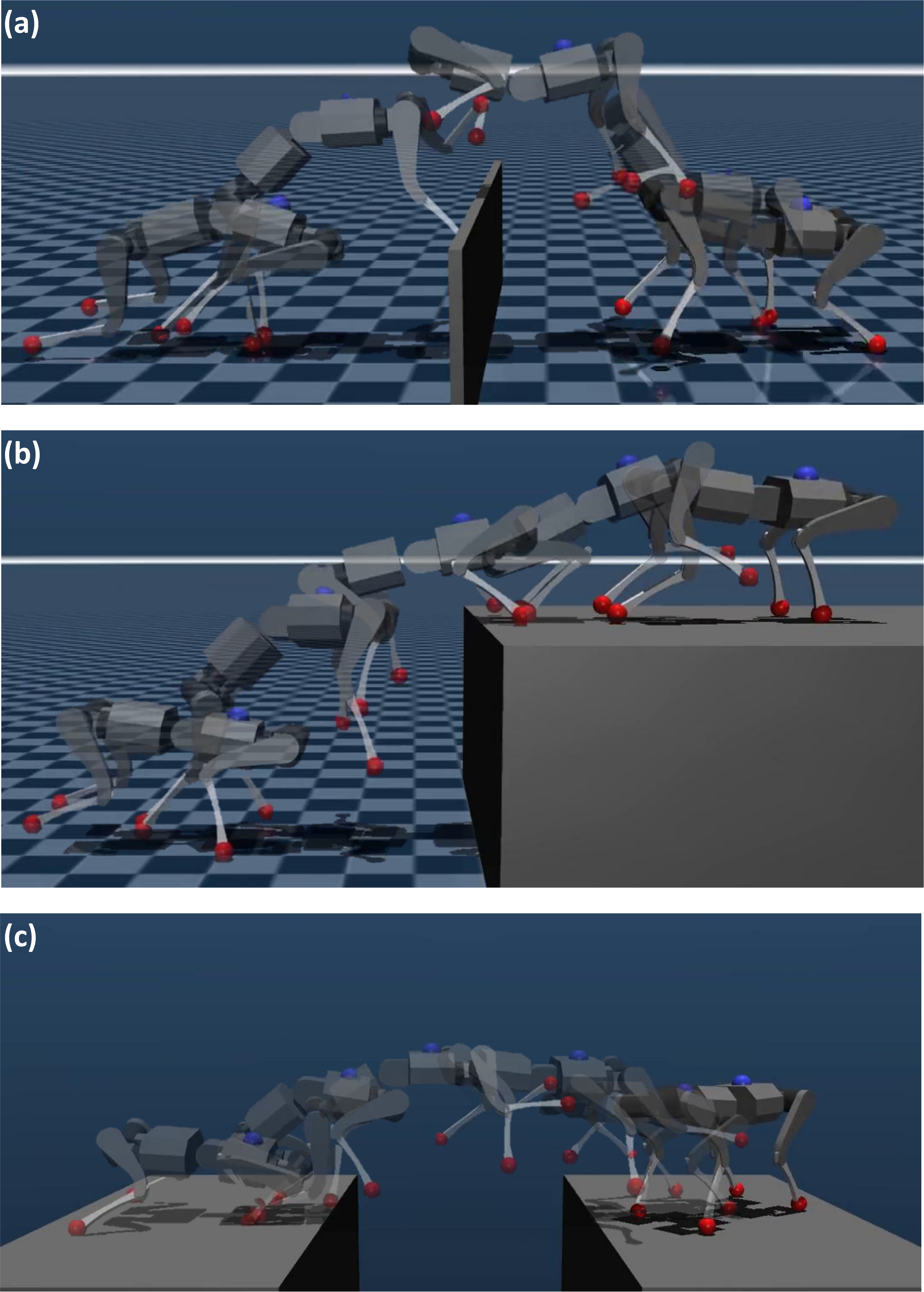}
\vspace{-1ex}
\caption{Snapshots of the learned policy executed on the MuJoCo simulator. (a) HurdleJump, (b) LedgeJump, (c) GapLeap}
\label{fig:sim2sim}
\vspace{-3ex}
\end{figure}

Snapshots of the HurdleJump, LedgeJump, and GapLeap motions executed on the MuJoCo \cite{todorov2012mujoco} simulator are shown in \figref{fig:sim2sim}.

The learned policy successfully reproduces similar motions on the MuJoCo simulator.

\section{Discussion}\label{sec:discussion}

\paragraph{Advantages and Challenges of Wrench-Based Exploration}
We experimentally investigated how introducing a wrench into the action space affects learning characteristics. The advantages and challenges are summarized as follows.

The first advantage is accelerated learning. By including a wrench in the action space, the agent obtains higher exploration capability compared to joint-space control alone. This expands the range of exploration that leads to the discovery of target motions and shortens the time required to acquire them.

The second advantage is the reduction of heuristic learning design.
When learning only with joint actions, the limited exploration capability often requires artificial adjustments such as detailed reward tuning and step-by-step curricula. Introducing a wrench broadens the exploration space, allowing learning to progress with less reliance on such auxiliary designs.
\paragraph{Improvement of the Learning Framework}
For simplicity, the wrench policy and joint policy were trained with the same reward function, although their roles are inherently different. The primary role of the wrench policy is to facilitate learning of the joint policy.

Using the same reward for both policies may cause the system to favor a wrench policy that does not sufficiently consider embodiment.
To address this issue, the reward for the wrench policy should instead evaluate how effectively it promotes the autonomous achievement of the target motion by the joint policy.
One possible approach is to introduce a structure in which joint policy learning forms an inner loop while its progress is optimized externally, similar to hierarchical reinforcement learning.

There is also room for improvement on the joint-policy side. The current framework adapts to a gradually decaying wrench, but another approach is to use wrench-generated motions as an imitation target.
For example, trajectories obtained from wrench-driven motions could be used as teacher signals, enabling the joint policy to acquire motions that are more consistent with the robot's embodiment.

\paragraph{Future Development Potential}
The idea of incorporating a wrench into the action space is not limited to a specific robot morphology.
Although this study focused on a quadruped robot, the approach may also be effective for robots with different locomotion forms, such as bipedal or multi-legged systems, particularly in tasks requiring dynamic control.

Furthermore, the wrench policy is relatively weakly tied to a specific body structure.
This abstraction suggests the possibility of transferring motion knowledge between different embodiments.
By using the wrench as a body-independent representation, it may be possible to construct a framework for sharing and transferring motor knowledge across robots.

From this perspective, the proposed approach contributes not only to improving learning efficiency but also to expanding the design principles of reinforcement-learning-based robot control.

\section{Conclusion}\label{sec:conclusion}

In this study, we proposed a new method, WARL, which introduces a wrench into the action space to overcome the constraints on exploration capability in reinforcement learning for legged robots.
The proposed method integrates wrench-guided exploration and success rate-based curriculum learning.

The experimental results demonstrated that by using WARL, it is possible to learn motions uniformly under various conditions without requiring individual design for each terrain or task.
In addition, an ablation study verified that the Switching Curriculum, which gradually removes the wrench, is effective for the autonomous acquisition of a joint policy.

On the other hand, it also became clear that the introduction of a wrench tends to lead to the learning of motions that do not fully utilize the robot's specific embodiment.
This issue is attributed to the fact that the wrench is an abstract input that is relatively independent of the body structure.

In the future, further refinement of the learning framework is necessary, such as a method for generating wrenches that explicitly considers embodiment and a transition from exploration using wrenches to the acquisition of a joint control policy.

{
  \bibliographystyle{IEEEtran}
  \bibliography{bib}
}

\end{document}